\documentclass[10pt, conference, compsocconf]{IEEEtran}
\IEEEoverridecommandlockouts

\usepackage{cite}
\usepackage{amsmath,amssymb,amsfonts}
\usepackage{algorithmic}
\usepackage{graphicx}
\usepackage{textcomp}
\usepackage{xcolor}

\usepackage{booktabs} 
\graphicspath{ {figure/} } 
\usepackage{subfig}
\usepackage[linesnumbered,ruled]{algorithm2e}
\usepackage{verbatim}
\usepackage{enumerate}
\usepackage{setspace}

\begin{document}

\title{CN-Probase: A Data-driven Approach for Large-scale Chinese Taxonomy Construction}

\author{\IEEEauthorblockN{Jindong Chen\textsuperscript{1}, 
		Ao Wang\textsuperscript{1}, 
		Jiangjie Chen\textsuperscript{1}, 
		Yanghua Xiao\textsuperscript{12*}\thanks{ *Yanghua Xiao is the corresponding author. This paper is the was supported by National Key R\&D Program of China (No. 2017YFC0803700), by National NSFC (No.61732004) and by Shanghai Municipal Science and Technology project (No.16JC1420400).},
		Zhendong Chu\textsuperscript{1}, 
		Jingping Liu\textsuperscript{1}, 
		Jiaqing Liang\textsuperscript{13}, 
		Wei Wang\textsuperscript{1}}
\IEEEauthorblockA{
\textsuperscript{\rm 1}Shanghai Key Laboratory of Data Science, School of Computer Science, Fudan University, China\\
\textsuperscript{\rm 2}Shanghai Institute of Intelligent Electronics \& Systems, Shanghai, China\\
\textsuperscript{\rm 3}Shuyan Technology, Shanghai, China\\
\{chenjd16, awang15, jiangjiechen14, shawyh, zdchu15, jpliu17\}@fudan.edu.cn, l.j.q.light@gmail.com, weiwang1@fudan.edu.cn}
}

\maketitle

\begin{abstract}
	Taxonomies play an important role in machine intelligence. However, most well-known taxonomies are in English, and non-English taxonomies, especially Chinese ones, are still very rare. In this paper, we focus on automatic Chinese taxonomy construction and propose an effective generation and verification framework to build a large-scale and high-quality Chinese taxonomy. In the generation module, we extract \textit{isA} relations from multiple sources of Chinese encyclopedia, which ensures the coverage.  To further improve the precision of taxonomy, we apply three heuristic approaches in verification module. As a result, we construct the largest Chinese taxonomy with high precision about 95\% called CN-Probase. Our taxonomy has been deployed on Aliyun, with over 82 million API calls in six months.
\end{abstract}


\section{Introduction}
Semantic networks and conceptual taxonomies are playing an increasingly important role in many applications.
Conceptual taxonomies are composed of entities, concepts and hypernym-hyponym relations (a.k.a \textit{isA} relations).
For example, \texttt{apple} \textit{isA} \texttt{fruit}, where \texttt{fruit} is the \emph{hypernym} of \texttt{apple}.
The opposite term for hypernym is \emph{hyponym}, so \texttt{apple} is the hyponym of \texttt{fruit}.
We use the expression \textit{isA}(\texttt{A}, \texttt{B}) to express a hypernym-hyponym relationship, which means \texttt{A} is a hyponym of \texttt{B}.

\begin{figure}[!hbt]
	\centering
	\includegraphics[width=0.90\linewidth]{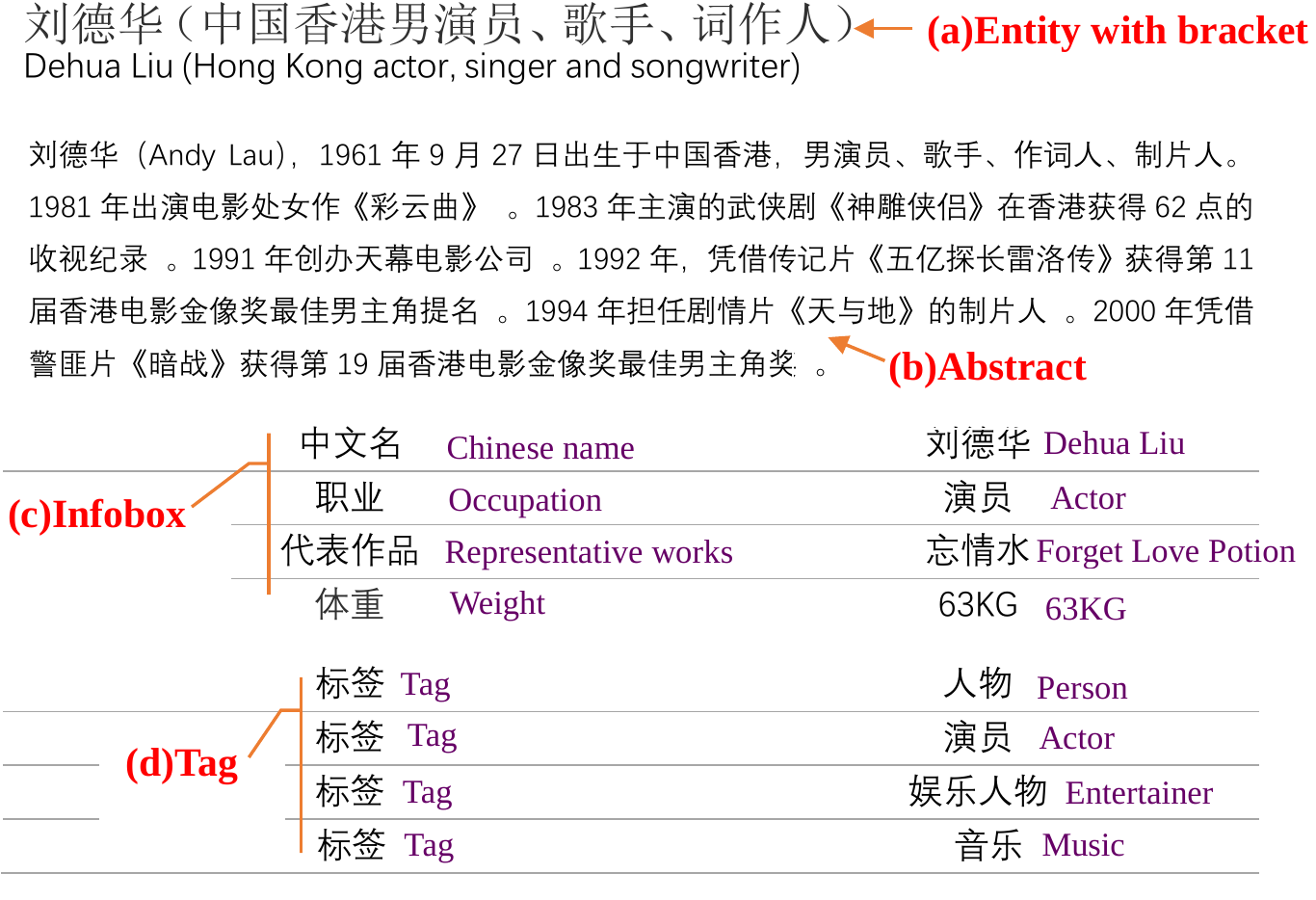}
	\caption{Page in Chinese encyclopedia. }
	\label{fig:cndbpedia}
\end{figure}

Early taxonomies such as WordNet \cite{Miller1995WordNet} and Cyc \cite{Lenat1989Building} are built by human experts. They are highly precise but limited in \emph{coverage} and are expensive to construct. 
Therefore, most of succeeding research efforts are devoted to constructing taxonomies \emph{automatically} from web corpus or online encyclopedia.
These efforts have produced a lot of well-known English taxonomies such as WikiTaxonomy \cite{Ponzetto2008WikiTaxonomy} and Probase \cite{wu2012probase}. 
However, non-English taxonomies, especially Chinese ones, are still very rare. The direct reason is the complexity of Chinese. Chinese is a lower-resourced language with flexible expressions and grammatical rules \cite{Li2015User}.
For example, the order of words in Chinese can be changed flexibly in the sentence. Besides, Chinese has no word spaces, no explicit tenses and voices, no distinct singular/plural forms.

In this paper, we aim at constructing a large-scale and high-quality Chinese taxonomy automatically from a Chinese encyclopedia website. 
Our observation is that there are multiple sources including bracket, abstract, infobox, tag in Chinese encyclopedia marked as (a), (b), (c), and (d) respectively, as shown in Figure \ref{fig:cndbpedia}. We highlight that these information, yet to be fully leveraged from previous work, is the key to achieve our objective. 
For example, the bracket in Figure 1 allows us to extract \textit{\textit{isA}}(\texttt{Dehua Liu}, \texttt{singer}).  The triples in infobox, such as $<$\texttt{Dehua Liu, occupation, actor}$>$, allows us to extract \textit{\textit{isA}}(\texttt{Dehua Liu}, \texttt{actor}).
And some tags directly tell us that \textit{\textit{isA}}(\texttt{Dehua Liu}, \texttt{person}).
Thus, the full usage of these information allows us to find a significant number of \textit{isA} relations.
However, the tags still contain noise, and the inference of hypernyms from triples and text are still error-prone. For example, \textit{\textit{isA}}(\texttt{Dehua Liu}, \texttt{music}) is a wrong \textit{isA} pair extracted from tag.

\begin{figure*}[!hbt]
	\centering
	\includegraphics[width=0.90\linewidth]{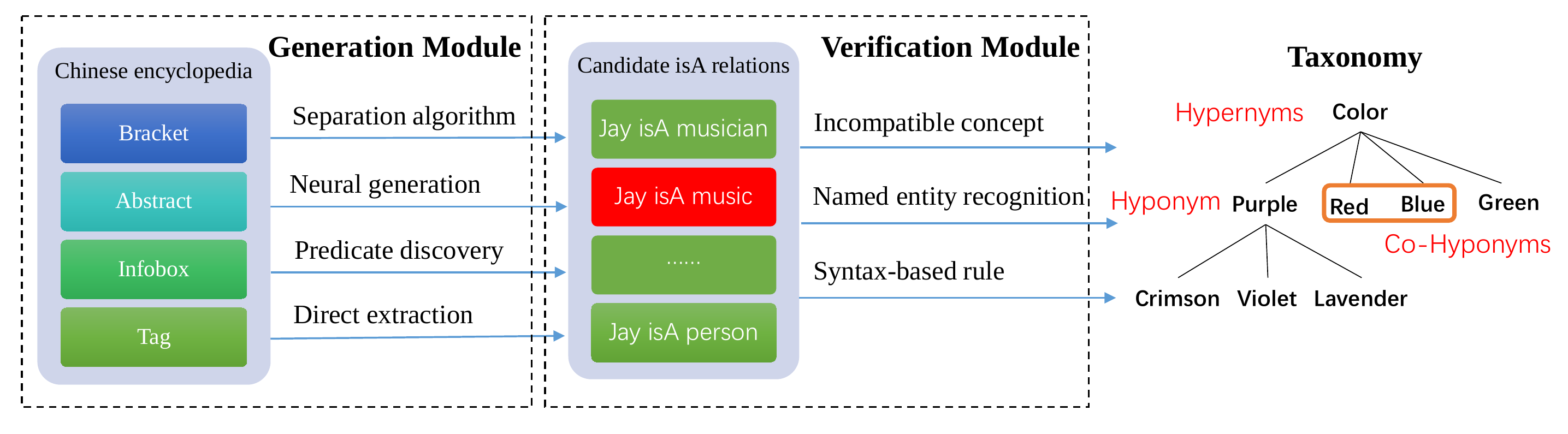}
	\caption{Framework of CN-Probase.}
	\label{fig:framework}
\end{figure*}

To solve the above problems, we propose a \emph{generation and verification} framework, which is shown in Figure \ref{fig:framework}. 
The input of our framework is Chinese encyclopedia. In the \textit{generation step}, we leverage different algorithms to extract \textit{isA} relations from multiple sources of Chinese encyclopedia, which ensures the \emph{coverage}.
The candidate \textit{isA} relations are produced by merging all \textit{isA} relations generated from different sources of Chinese encyclopedia.
In the \textit{verification step}, we employ three heuristic strategies to remove noise, which ensures the \textit{precision} of \textit{isA} relations. A candidate \textit{isA} relation will be filtered if any of the strategies makes the judgment that it is a wrong case. Our contributions in this paper can be summarized as follows.

\begin{itemize}
	\item We design an effective generation and verification framework for Chinese taxonomy construction. 
	\item We build the large-scale Chinese conceptual taxonomy CN-Probase with high precision (95\%), 
	including 15 million disambiguated entities, 270 thousand distinct concepts and 33 million \textit{isA} relations.
	\item In experiments, we demonstrate the size, precision and coverage of CN-Probase.
\end{itemize}

\section{Generation Module} \label{gm}
In this section, we acquire \textit{isA} relations from four sources of Chinese encyclopedia (i.e., \textit{bracket, abstract, infobox} and \textit{tag}) by four corresponding algorithms.

\begin{figure}[!hbt]
	\centering
	\includegraphics[width=0.9\linewidth]{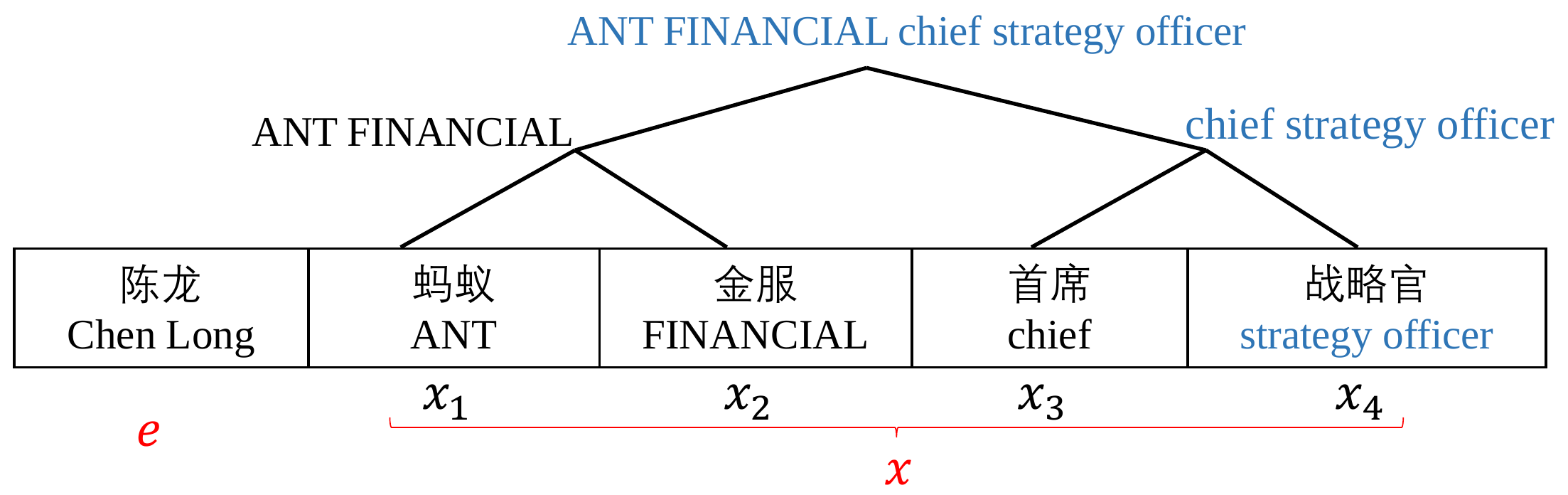}
	\caption{Example of hypernym acquisition. The blue phrases are the hypernyms of the entity.}
	\label{fig:head_modifier1}
\end{figure}

\textit{Separation algorithm} is proposed to acquire the hypernyms of the entity from the noun compound in the bracket. The input of the algorithm is a disambiguated entity denoted as $e(x)$, where $e$ is the entity name and $x$ is the noun compound. Let $(x_1, x_2,...,x_n)$ be the word sequence of length $n$ by conducting word segmentation on $x$. An example is shown in Figure \ref{fig:head_modifier1}, \texttt{ANT FINANCIAL chief strategy officer} is segmented into \{\texttt{ANT}, \texttt{FINANCIAL}, \texttt{chief}, \texttt{strategy officer}\}. The output of the algorithm is the hypernyms of the input entity. Let $\oplus$ denote the operation of string concatenation. The algorithm begins with the rightmost three elements of the word sequence. For simplicity, we will use $(x_{i-1}, x_i, x_{i+1})$ to explain the algorithm:
\begin{description}
	\item[Step 1:]\ Given $(x_{i-1}, x_i, x_{i+1})$, if PMI($x_{i-1}, x_i$)$<$PMI($x_i$, $x_{i+1}$) holds, the algorithm goes to \textit{step 2}, otherwise goes to \textit{step 3}.
	\item[Step 2:]\ Separate the sequence as $(x_{i-1}, x_i\oplus x_{i+1})$. Then move the sliding window to left by one unit (word) and acquire $(x_{i-2}, x_{i-1}, x_i\oplus x_{i+1})$, go to \textit{step 1}.
	\item[Step 3:]\ Move the sliding window to left by one unit and acquire $(x_{i-2}, x_{i-1}, x_i)$, then go to \textit{step 1}. 
	\item[Step 4:]\ When the leftmost element $x_1$ locates in sliding window and the sequence $(x_1,x_2,x_3)$ satisfies PMI$(x_1,x_2)$$>$PMI$(x_2,x_3)$, we separate it as $(x_1\oplus x_2,x_3)$. Move the sliding window to right by one unit and acquire $(x_1\oplus x_2,x_3,x_4)$, then go to \textit{step 1}.
\end{description}
The output of the algorithm is a binary tree. We extract all the leaf nodes along with the rightmost path of the binary tree as the hypernyms. To the best of our knowledge, we are the first to extract \textit{isA} relations from entity brackets, and we obtain nearly 2 million \textit{isA} relation with a precision of 96.2\% from this data source.

\textit{Neural generation} is used to obtain the hypernym (concept) of an entity from the abstract of the entity. We first utilize \textit{distant supervision} \cite{Mintz2009Distant} to construct dataset $\{(x_1,y_1), (x_2, y_2),..., (x_n, y_n)\}$ where $n$ is the number of samples. 
We have acquired numerous \textit{isA} relations with a precision over 96\% from bracket . 
For $i$-th \textit{isA} relation, we regard the abstract of its hyponym as $x_i$ and the hypernym as $y_i$.
In this way, we build the dataset consisting of more than 300,000 samples. 
Then, we employ an encoder-decoder model  to generate concepts from the abstract. But merely using this basic model suffers from out-of-vocabulary (OOV) problem.
Hence we use copynet\cite{gu2016incorporating} to perform this task.

\textit{Predicate discovery} is proposed to acquire \textit{isA} relations from the infobox.
First, we apply the idea of \textit{distant supervision} \cite{Mintz2009Distant} to discover the predicates such as \texttt{occupation} which are the implicit \textit{isA} relationships.
Specifically, we employ the \textit{isA} relations that have been extracted from bracket as prior knowledge, since they have a precision over 96\%.
Then, we use these \textit{isA} relations (e.g., \textit{isA}(\texttt{Jay Chou}, \texttt{singer})) to align the SPO triples (e.g., $<$\texttt{Jay Chou}, \texttt{occupation}, \texttt{singer}$>$) and discover 341 candidate predicates (e.g., \texttt{occupation}) in total.
However, there are noises in these candidates. To further purify these candidates, we manually select 12 predicates as the implicit \textit{isA} relationships to acquire \textit{isA} relations from its corresponding SPO triples. 

\textit{Direct extraction} is used to obtain \textit{isA} relations from tag. A tag is a word or phrase which is used to describe the entities in Chinese encyclopedia. A majority of tags are the hypernyms of the entities. We directly regard the tags as the hypernyms of an entity.

\section{Verification Module} 
In this section, we propose three effective heuristic strategies to filter the wrong \textit{isA} relations produced in the generation module and improve the precision.

\subsection{Incompatible concepts}
Two concepts such as \texttt{singer} and \texttt{actor} are \textit{compatible} since they have some common entities. In some case, two concepts such as \texttt{person} and \texttt{book} sharing no entities are \textit{incompatible}, which motivates us to filter wrong \textit{isA} relations by detecting incompatible concept pairs.
Our approach is composed of two steps: incompatible concept pairs construction and wrong \textit{isA} relations detection. In the first step, we construct the incompatible concept pairs based on the Jaccard similarity between the hyponyms set of two concepts and the cosine simiarity between the distribution of concept attribute.
In the second step, given an entity $e$ and its two incompatible concepts $c_1$ and $c_2$, we detect the wrong one by the KL divergence:
\begin{equation}
D_{KL}(v_{att(e)}||v_{att(c)})=-\sum_{x}v_{att(e)}log\frac{v_{att(c)}}{v_{att(e)}}
\end{equation}
$v_{att(e)}$  and $v_{att(c)}$ are the attribute distribution of the entity $e$ and concept $c$. Then, we filter the concept with a larger KL score. 

\subsection{Named entity recognition} 
The fact that whether a hypernym is a \textit{named entity} (NE) plays an important role in detecting wrong \textit{isA} relations, since NE usually cannot be a hypernym of an entity.
For example, \textit{isA}(\texttt{iPhone}, \texttt{America}) is a wrong \textit{isA} relation due to the NE hypernym \texttt{America}. Inspired by the above observation, we detect wrong \textit{isA} relations by recognizing the NE hypernyms. We use $s_1(H)$ and $s_2(H)$ to represent the support of a hypernym $H$ as a NE in the Chinese text corpus and our taxonomy respectively. Specifically, $s_1(H)=NE(H)/total(H)$ where $NE(H), total(H)$ represent the number of occurrence of $H$ as a NE and the total number of occurrence of $H$ in Chinese text corpus respectively. Similarly, we acquire $s_2(H)$ by replacing Chinese text corpus with our taxonomy.
We further use a \textit{noisy-or model} to combine the two scores. 
\begin{equation}  
s(H)=1-(1-s_1(H))\cdot(1-s_2(H))
\end{equation} 
The rationale of the noisy-or model is to amplify the support signal. We set the threshold emperically and filter the \textit{isA} relations whose support $s(H)$ is greater than the threshold.

\subsection{syntax-based rule} We also use some syntax rules to further filter wrong \textit{isA} relations.
We describe the most typical rules as follows: (1) A good hypernym should not be a thematic word such as \texttt{politics}, \texttt{military}. We collect a Chinese lexicon from Li et al. \cite{Li2015User} including 184 non-taxonomies, thematic words. Then we filter the \textit{isA} relations whose hypernym in this lexicon;
(2) The stem of the lexical head of hypernym should not occur in the non-head position of the hyponym.
We filter the wrong candidate \textit{isA} relations \textit{isA}(\texttt{educational institution}, \texttt{education}) by this rule.

\section{EXPERIMENTS}\label{exp}
We apply the proposed framework on Chinese encyclopedia. As a result, we construct the large-scale and high-quality Chinese taxonomy including 270,026 distinct concepts, 15,066,667 disambiguated entities, 32,398,018 entity-concept relations and 527,288 subconcept-concept relations (32,925,306 \textit{isA} relations in total). Readers can refer to \textit{http://kw.fudan.edu.cn/cnprobase/search/} for complete experimental results.

\subsection{Experiment Settings} \label{es}
\subsubsection*{Data Source} CN-DBpedia \cite{Xu2017CN} is one of the largest open-domain Chinese encyclopedia derived from Baidu Baike Hudong Baike and Chinese Wikipedia. The experimental dataset is from CN-DBpedia dump generated on May 20, 2017, which includes 15,990,349 entities, 8,096,835 pieces of abstract information, 132,435,632 SPO triples and 19,929,407 tags.

\subsubsection*{Metrics}
There are five commonly used metrics in taxonomy evaluation: the number of entities, concepts and \textit{isA} relations, precision and coverage. To estimate the precision of taxonomies, we randomly select 2000 \textit{isA} relations in total from taxonomies and manually label whether a relation is correct or not.

\begin{table*}[!hbt]
	\center
	\caption{Comparisons with other taxonomies. `-' represents results that are not provided.}
	\begin{tabular}{|c|c|c|c|c|}
		\hline
		Taxonomy&\# of entities&\# of concepts&\# of \textit{isA} relations & precision\\
		\hline
		Chinese WikiTaxonomy&581,616&79,470&1,317,956&97.6\% \\ 
		Bigcilin&9,000,000&70,000&10,000,000& 90.0\% \\ 
		Probase-Tran&404,910&151,933&1,819,273&54.5\%\\ 
		\hline
		\textbf{CN-Probase}&\textbf{15,066,667}&270,025&\textbf{32,925,306}&95.0\%\\
		\hline
	\end{tabular}
	\label{table:taxonomy_comparision}
\end{table*}

\subsubsection*{Baselines} 
We compare CN-Probase with the following well-known Chinese taxonomies including Chinese WikiTaxonomy \cite{Li2015User} ,
Bigcilin \cite{fu2013exploiting} and Probase-Tran proposed by us. We translate Probase from English to Chinese by utilizing Google Translator. Then we use three heuristic methods from three aspects (meaning, transitivity, POS) to filter translation errors.

\subsection{Results}
The main results are shown in Table \ref{table:taxonomy_comparision}. Compared with other Chinese taxonomies, CN-Probase is the largest one when it comes to the number of entities, concepts and \textit{isA} relations. The main reason is that we extract \textit{isA} relations from multiple sources of Chinese encyclopedia.
Besides, CN-Probase is a high-quality taxonomy with a precision 95\% that outperforms Probase-Tran and Bigcilin. Although part of the noise has been reduced by three heuristic methods, the precision of Probase-Tran is still quite low due to various sources of noise.
Hence simple cross-language translation cannot produce high-quality Chinese taxonomy.
Bigcilin also extracts \textit{isA} relations from multiple sources, but its precision is worse than ours since we use verification module to further improve the precision.
Chinese WikiTaxonomy is built only from a single source (i.e. tag) of Chinese encyclopedia. As a result, it has a high precision but low coverage, and the number of \textit{isA} relations in our taxonomy is 25x larger than Chinese WikiTaxonomy. We also evaluate the precision of each source for our taxonomy, and the precision of \textit{isA} relations derived from the tag is 97.4\% which is comparable to Chinese WikiTaxonomy. 

Given that CN-Probase has more concepts and entities than other Chinese taxonomies such as Bigcilin and Probase-Tran, a reasonable question to ask is whether they are more effective in understanding text. We measure one aspect of the effectiveness here by examining CN-Probase's coverage on QA task. A question is said to be \textit{covered} by a taxonomy if the question contains at least one concept or entity within the taxonomy. The dataset is from the QA task of NLPCC2016 that includes 23,472 questions.
In all, CN-Probase covers 21,520 questions with a coverage of 91.68\%.
The covered entities have 2.14 concepts on average, 
which shows the significant effectiveness of CN-Probase in text understanding

\section{System and Application} \label{sa}
By adopting the proposed methods, CN-Probase has already been deployed on Aliyun.
We also publish three APIs on \textit{http://kw.fudan.edu.cn/apis/cnprobase/} to make our taxonomy accessible from Web. By September 2018, these APIs have already been called 82 million times by research institutions and companies since published on March 2018. Table \ref{table:api} shows the function of each API and their usage statistics.
So far, CN-Probase has been used in many applications including short text classification \cite{chen2019deep}, information extraction, etc.
\begin{table}[!hbt]
	\center
	\caption{APIs and their descriptions.}
	\begin{tabular}{|c|c|c|c|}
		\hline
		API name&Given&Return&Count   \\
		\hline
		men2ent & mention & entity &  43,896,044\\
		getConcept & entity & hypernym list & 13,815,076 \\
		getEntity & concept& hyponym list & 25,793,372 \\
		\hline
	\end{tabular}
	\label{table:api}
\end{table}

\section{Conclusion} \label{con}
In this paper, we propose an effective generation and verification framework for automatic taxonomy construction.
We build the large-scale and high-quality Chinese taxonomy CN-Probase from multiple sources of our Chinese encyclopedia. So far, CN-Probase has been used in many real-world applications.

\bibliographystyle{IEEEtran}
\setstretch{0}
\bibliography{sample-bibliography}

\end{document}